\documentclass{article}
\usepackage{ijcai24}

\usepackage{times}
\usepackage{soul}
\usepackage{url}
\usepackage[hidelinks]{hyperref}
\usepackage[utf8]{inputenc}
\usepackage[small]{caption}
\usepackage{graphicx}
\usepackage{amsmath}
\usepackage{amsthm}
\usepackage{booktabs}
\usepackage{algorithm}
\usepackage{algorithmic}
\usepackage[switch]{lineno}

\usepackage{multirow}
\usepackage{array}
\usepackage{subcaption}
\usepackage{tikz}
\usepackage{amssymb}
\usetikzlibrary{mindmap,trees,shadows,decorations.pathreplacing}

\title{From Words to Molecules: A Survey of Large Language Models in Chemistry}

\author{
Chang Liao$^1$
\and
Yemin Yu$^2$\and
Yu Mei$^3$\And
Ying Wei$^1$\\
\affiliations
$^1$School of Computer Science and Engineering, Nanyang Technological University, Singapore\\
$^2$City University of Hong Kong\\
$^3$College of Computer Science and Technology, Zhejiang University\\
\emails
chang019@e.ntu.edu.sg,
yeminyu2-c@my.cityu.edu.hk,
meiyu1997@zju.edu.cn,
ying.wei@ntu.edu.sg
}

\begin{document}

\maketitle

\begin{abstract}
In recent years, Large Language Models (LLMs) have achieved significant success in natural language processing (NLP) and various interdisciplinary areas. However, applying LLMs to chemistry is a complex task that requires specialized domain knowledge. This paper provides a thorough exploration of the nuanced methodologies employed in integrating LLMs into the field of chemistry, delving into the complexities and innovations at this interdisciplinary juncture. Specifically, our analysis begins with examining how molecular information is fed into LLMs through various representation and tokenization methods. We then categorize chemical LLMs into three distinct groups based on the domain and modality of their input data, and discuss approaches for integrating these inputs for LLMs. Furthermore, this paper delves into the pretraining objectives with adaptations to chemical LLMs. After that, we explore the diverse applications of LLMs in chemistry, including novel paradigms for their application in chemistry tasks.  Finally, we identify promising research directions, including further integration with chemical knowledge, advancements in continual learning, and improvements in model interpretability, paving the way for groundbreaking developments in the field.
\end{abstract}

\section{Introduction}

Humans understand and describe their environment using natural language, which reflects the complexity of human thought. The emergence of Large Language Models (LLMs) marks a significant advancement in artificial intelligence, showcasing remarkable abilities in various domains. These models excel at understanding and generating complex text, making them crucial for tasks that demand deep textual analysis and creation.

Intriguingly, a scientific domain such as chemistry has its unique language, akin to the way humans utilize natural languages. In chemical processes, bonds are broken, and atoms are exchanged during reactions, similar to how syntax operates, while molecules are formed within specific physical constraints, echoing the principles of grammar. This parallel suggests the potential for encoding chemical information into LLMs in a manner comparable to natural language. Despite the conceptual parallels, the languages of chemistry and human communication differ substantially in their semantics. Consequently, incorporating chemical knowledge into LLMs presents a complex challenge, with numerous approaches being explored to leverage LLMs in the field of chemistry, making it a subject of considerable interest.

\begin{figure}
    \centering
    \includegraphics[width=1\linewidth]{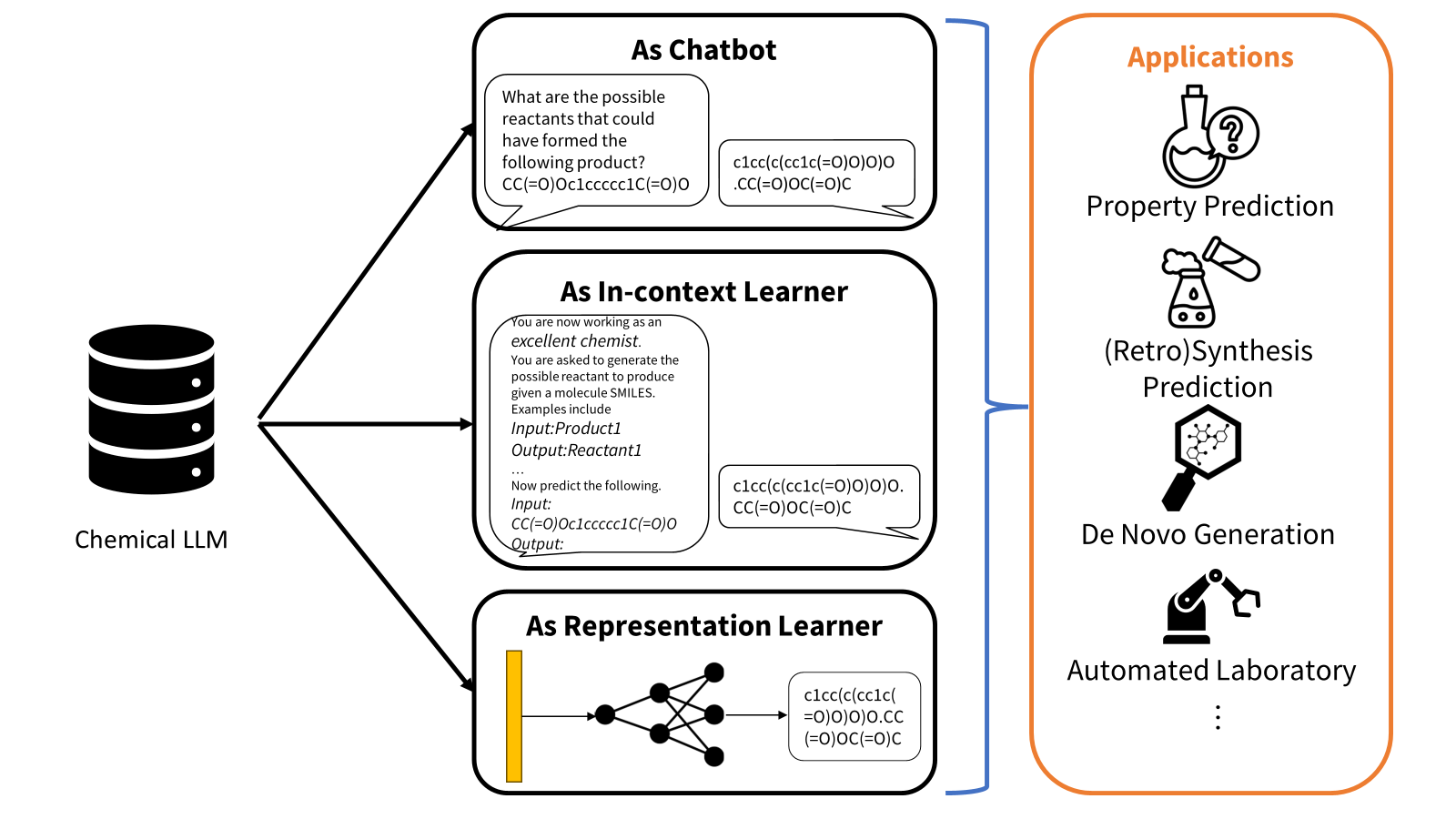}
    \caption{LLMs for Chemistry: Applications and Paradigms}
    \label{fig:llmsinchemistry}
\end{figure}

With numerous approaches being proposed, there's currently no systematic survey focused specifically on the application of Large Language Models (LLMs) in the field of chemistry. \cite{xia_systematic_2023} comes closest with a systematic survey on chemical pretrained models, encompassing LLMs along with pretrained models in other modalities such as graph and image. However, this survey primarily addresses the general objectives and utilization of pre-trained models and overlooks the nuanced application paradigms of LLMs, treating them primarily as representation learners. This perspective neglects the unique ways LLMs can be integrated into chemical research, as illustrated in Figure~\ref{fig:llmsinchemistry}. 

\begin{figure*}[h]
    \centering
    \includegraphics[width=0.9\linewidth]{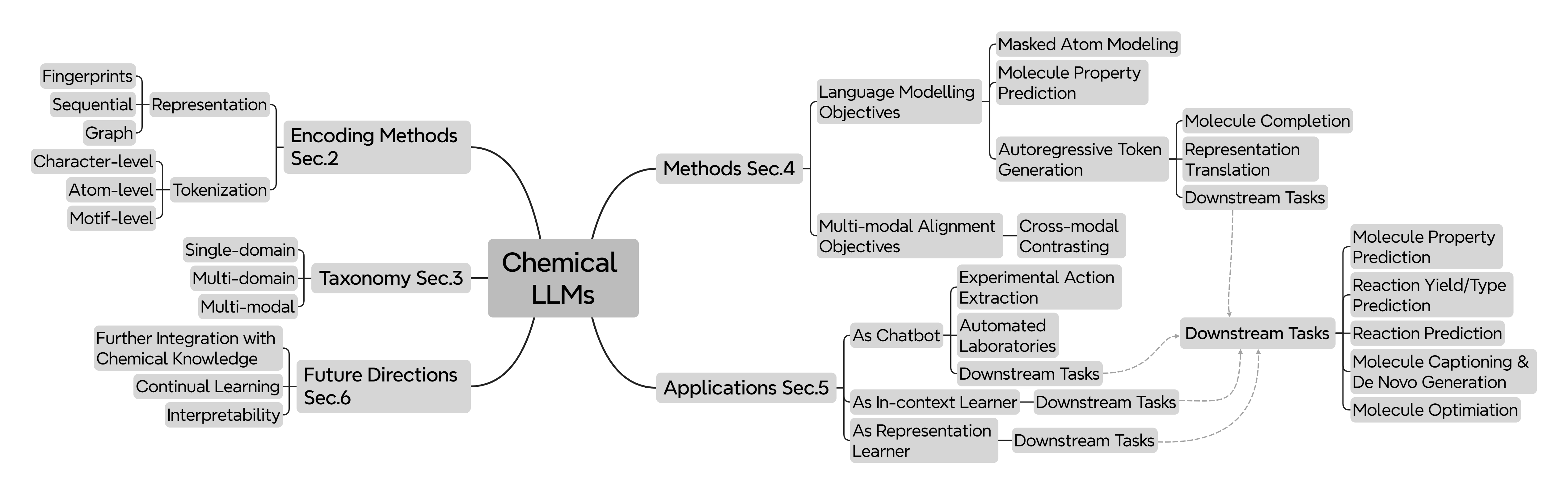}
    \caption{An overview of topics in this paper, with dash lines indicating their applicability to various downstream tasks.}
    \label{fig:overview}
\end{figure*}

To differentiate our work from \cite{xia_systematic_2023}, we summarize our contributions as follows:
\begin{enumerate}
\item We provide a comprehensive review of tokenization methods for molecular sequences, categorizing them based on their granularity.
\item We offer a systematic taxonomy of existing approaches, based on the nature of pretraining data, and discuss methodologies for adapting chemical data within LLM frameworks, including how to integrate chemical data with other domains or modalities to enhance LLM performance.
\item We investigate the nuances of applying self-supervised learning on chemical data, highlighting domain-specific opportunities and examining tailored techniques for chemical tasks.
\item We identify unique paradigms for LLM utilization in chemistry, presenting applications exclusive to their capabilities and elucidating novel contributions to chemical research.
\item We outline several promising future research directions, exploring emerging trends in both chemistry and LLM development that could significantly advance the interdisciplinary field of chemical LLMs.
\end{enumerate}

The structure of this survey is depicted in Figure~\ref{fig:overview}, serving as a guide for readers throughout this paper.



\section{Molecule encoding methods}
\label{sec:mem}

In order for LLMs to learn from molecules, molecules must be represented in a series of discrete tokens. In this section, we will provide a concise overview of contemporary methods for molecular representation and tokenization. 

\subsection{Representations Methods of Molecules}
\paragraph{Fingerprint Representations}
Molecular fingerprints are typically represented as a binary string (a series of 0s and 1s), where each position in the string (bit) corresponds to a particular structural feature or property of the molecule. For instance, one bit might represent the presence or absence of a certain chemical group. There are several types of fingerprints, each capturing different aspects of molecular structure like molecular access system (MACCS) keys  \cite{durant_reoptimization_2002} and ECFP (Extended-Connectivity Fingerprints)  \cite{rogers_extended-connectivity_2010}.

\paragraph{Sequential Representations}
Simplified Molecular-Input Line-Entry System (SMILES)  \cite{weininger_smiles_1988} is the first sequential molecular representation, it is compact and human-readable. However, SMILES suffers from (1) \textit{non-uniqueness}, as a single molecule could be represented by multiple valid SMILES strings, (2) \textit{non-robustness}, as SMILES strings do not inherently ensure chemically feasible structures, (3) \textit{information-loss}, as SMILES doesn't explicitly convey structural information. 
Several innovations have been proposed to address those limitations.
SELFIES \cite{krenn_self-referencing_2020} ensures robustness with strict derivation rules. International Union of Pure and Applied Chemistry (IUPAC) Chemical Identifier (InChI) \cite{heller_inchi_2013} focuses on uniqueness through a complex hierarchical representation convention. 

\paragraph{Graph Representations}
Molecular graph representations, crucial in cheminformatics and drug design, vary from two-dimensional to high-dimensional forms. Two-dimensional (2D) types, such as molecular fingerprints (ECFP) \cite{rogers_extended-connectivity_2010}, condense molecular structures into vectors for simpler similarity analysis but may overlook complex conformations. To compensate, high-dimensional representations, including three-dimensional (3D) details, have been developed. These employ a 3D coordinate system to accurately represent molecular structures. 
Further, four-dimensional (4D) molecular graph representations \cite{hopfinger1997construction} capture weighted utilization of diverse spatial configurations of molecules, enhancing understanding of molecular structures and interactions.

\subsection{Tokenization Methods of Molecules}

\begin{figure*}[h]
    \centering
    \includegraphics[width=0.9\textwidth]{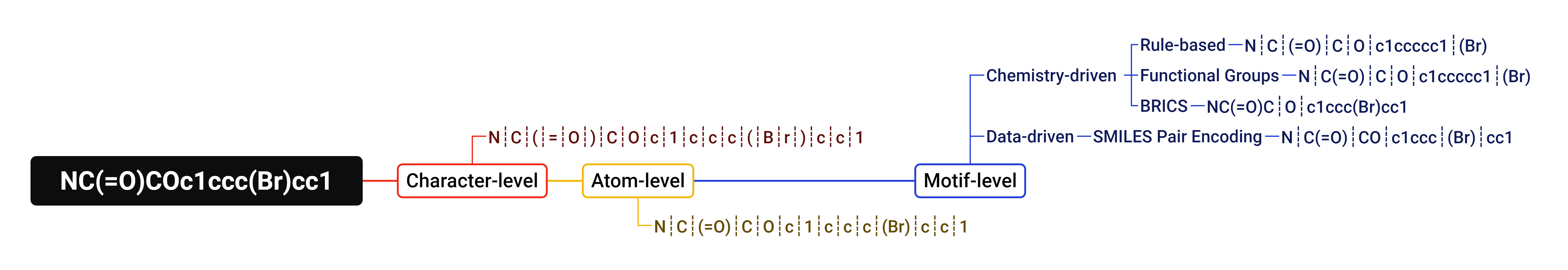}
    \caption{An Example of Tokenized Output from Different Tokenizers for the Sequence ``NC(=O)COc1ccc(Br)cc1''}
    \label{fig:Tokenizers}
\end{figure*}

The tokenization of molecule sequences can be primarily categorized into three lines of approaches: (1) \textbf{character-level}, (2) \textbf{atom-level}, and (3) \textbf{motif-level}.

\textbf{Character-level} tokenization treats each character as a separate token, leading to erroneous splitting of multi-character entities like ‘Br’. However, despite implausibility, this approach has shown effectiveness in chemical LLMs, as demonstrated in \cite{wang_smiles-bert_2019,edwards_translation_2022,lu_unified_2022,winter_smile_2022}, underscoring LLMs' impressive comprehension capabilities.

\textbf{Atom-level tokenization methods} offer a more rational approach by segmenting sequences into atoms. Recent sequential representation methods have introduced customized atom-level tokenizers, as seen in \cite{heller_inchi_2013,krenn_self-referencing_2020}. These in-house tokenizers are tailored to efficiently manage unique characters, such as ``[Ring1]'' in SELFIES and ``/h'' in InChI, which are essential for representing complex chemical structures within their respective formats. In contrast, SMILES does not come with a built-in tokenizer and is typically tokenized using regex-based expressions, as detailed in \cite{schwaller_found_2017}.

\textbf{Motif-level tokenization methods} can be done with \textit{chemistry-driven} approaches or \textit{data-driven} approaches.
\textit{Chemistry-driven} approaches involve breaking molecules into chemically meaningful substructures with the help of expert chemical knowledge. These methods are prominently featured in graph-related studies  \cite{jin_junction_2019,rong_self-supervised_2020,zhang_motif-based_2021}. For instance,  \cite{rong_self-supervised_2020} matches molecules to a database of functional groups,  \cite{jin_junction_2019} forms motifs by applying customized fragmentation rules to break bonds, and \cite{zhang_motif-based_2021} combines breaking of retrosynthetically interesting chemical substructures (BRICS)  \cite{degen_art_2008} with customized fragmentation rules to achieve a finer granularity.
Following those methodologies, \cite{feng_unimap_2023,xie_self-supervised_2023} have performed motif-level tokenization on sequential molecular representations. This involves breaking down molecules in the graph domain and transforming the fragmented graph motifs to sequential representations. Although these approaches produce chemically sound substructures, defining customized rules requires expert knowledge. Moreover, the process of sequence-graph transformation and matching predefined patterns like functional groups or BRICS, can be costly.  

\textit{Data-driven} approaches are inspired by subword-level tokenization methods like byte-pair encoding (BPE) \cite{gage_new_1994} in natural language processing, which iteratively merges the most frequent pairs of characters into a single token. A similar tokenization method was proposed by  \cite{li_smiles_2021} on SMILES. Other works have also adopted this subword-level tokenization by directly applying BPE on molecular sequential representation corpora  \cite{chithrananda_chemberta_2020,zhu_featurizations_2022,ahmad_chemberta-2_2022,xue_x-mol_2022,chilingaryan_bartsmiles_2022,christofidellis_unifying_2023,li_druggpt_2023,liu_multi-modal_2023,liu_git-mol_2023}. The methodology of uncovering common subpatterns in subword tokenization is akin to the discovery of motifs in chemistry, justifies the classification of this method as motif-level.

For illustrative examples of these various tokenization methods, please refer to Figure~\ref{fig:Tokenizers}, which depicts a comprehensive comparison of previously mentioned tokenizers on an example SMILES string NC(=O)COc1ccc(Br)cc1.

\section{Taxonomy}
\label{sec:tax}

We categorize current methodologies into three distinct groups by the knowledge within the pretraining corpus:
\begin{enumerate}
    \item \textbf{Single-domain} approaches pretrain their models purely on tokens from the chemical domain, i.e., sequential molecular representations and/or molecular properties tokens, without any tokens from the common text domain. Datasets containing vast amounts of molecules like PubChem \cite{kim_pubchem_2016} are adopted for this category.
    This kind of approach is mostly for representation learning and overlaps with pretrained models reviewed in \cite{xia_systematic_2023}, but more insights will be provided in the following sections for readers to appreciate the nuances of applying LLMs in the chemistry domain.
    
    \item \textbf{Multi-Domain Approaches}  entail pretraining on a corpus or corpora that merges chemical and common textual tokens. The Colossal Clean Crawled Corpus (C4) \cite{c4} is widely recognized for general text pretraining. Additionally, datasets featuring textual molecular descriptions aid in connecting chemical and text tokens. Notable datasets in this category include ChEBI-20 \cite{edwards_text2mol_2021}, PCDes \cite{zeng_deep-learning_2022}, PubChemSTM \cite{liu_multi-modal_2023}, PubChem324k \cite{liu_molca_2023}, and MoMu \cite{su_molecular_2022}. Embedding specific chemical tasks within text sequences further enhances multi-domain pretraining. Mol-Instructions \cite{fang_mol-instructions_2023} exemplifies this, offering a rich dataset for multi-domain pretraining with sentences crafted for tasks such as property prediction, reaction prediction, and also molecule description.
    
    There are multiple ways to mix tokens from different domains.
    MolT5 \cite{edwards_translation_2022} pretrains on sequences from different domains in one mini-batch. For instance, it processes a chemical sequence like ``ON=CCC1=c[NH1]C2=CC=CC=C12'' alongside a common text sequence such as ``Lissamine fast yellow(2-) is an organosulfonate oxoanion resulting from the removal of a proton.'' in one batch and perform pretraining objectives on them separately.
    

    A more common approach is to ``wrap'' tokens from different domains together in one sentence. 
    For example, ``Acetylsalicylic acid CC(=O)Oc1ccccc1C(=O)O appears as odorless white crystals or crystalline powder with a slightly bitter taste.'' contains both chemical sequence ``CC(=O)Oc1ccccc1C(=O)O'' and text sequence describing its properties. 
    However, a notable challenge arises from this method: identical tokens from different domains can represent distinct entities. For example, the character ``C'' might denote a carbon atom in molecular sequences, cysteine in protein structures, or simply the letter ``c'' in text. To resolve this ambiguity, specific tokens are often employed to delineate sequential representations, marking their beginning and end like [START\_SMILES] and [END\_SMILES]. Additionally, domain-specific indicators such as [SMILES\_C], [PROTEINS\_C], and [C] can be added to the vocabulary to clearly differentiate the context of each token.
 
    \item \textbf{Multi-modal} approaches advance further by incorporating information from various modalities into LLM. Molecular fingerprints, molecular graphs, and their corresponding images are frequently utilized alongside general text and molecular sequences. Each data modality is processed via specialized encoders, such as Transformer \cite{vaswani2017attention} for fingerprints, GIN \cite{xu2019powerful} for 2D graphs, SchNet \cite{schutt2017schnet} for 3D graphs, and ResNet \cite{he2016deep} for images. Regarding the corresponding training data in different modalities, the previously mentioned molecular dataset, such as \cite{irwin_zinc20free_2020}, can be used to retrieve molecular sequences and generate data in other modalities like graphs, fingerprints, and images with the help of cheminformatics tools like RDKit. Apart from being generated with cheminformatics tools, graph data can also be retrieved from datasets such as GEOM-Drugs \cite{axelrod2022geom}. Text encoders can be pretrained on datasets that include molecular descriptions like ChEBI-20 \cite{edwards_text2mol_2021}. After pretraining on different modalities, several adaptors or cross-modal attention layers are employed to align their latent spaces with that of LLMs for integration. This alignment presents a significant challenge, which will be thoroughly examined in the following sections.

    \end{enumerate}


\section{Methods}
\label{sec:methods}

This section delves into current pretraining methodologies, contrasting single-modal with multi-modal objectives. A detailed review of these methods is provided in Table~\ref{tab:methods}.

\begin{table*}[]
\centering
\resizebox{\textwidth}{!}{%
\begin{tabular}{@{}lccccccc@{}}
 &
  \textbf{Method} &
  \textbf{Backbone} &
  \textbf{Input Representation} &
  \textbf{Tokenization Methods} &
  \textbf{Pretraining Objectives} &
  \textbf{Applications} &
  \textbf{Reference} \\ \midrule
\multirow{15}{*}{\rotatebox[origin=c]{90}{\textbf{Single-Domain}}} &
  SMILES-BERT &
  BERT &
  SMILES &
  Character-level &
  MLM &
  Molecule Property Prediction & \cite{wang_smiles-bert_2019}
   \\\cmidrule(l){2-8}
 &
  Molecular Transformer &
  \begin{tabular}[c]{@{}c@{}}Autoregressive\\ Encoder-decoder\end{tabular} &
  SMILES &
  Atom-level &
  \begin{tabular}[c]{@{}c@{}}ATG\\ (Reaction Prediction)\end{tabular} &
  Reaction Prediction & \cite{schwaller_molecular_2019}
   \\\cmidrule(l){2-8}
 &
  ChemBERTa &
  RoBERTa &
  SMILES/SELFIES &
  Atom-level &
  MLM &
  Molecule Property Prediction & \cite{chithrananda_chemberta_2020}
   \\\cmidrule(l){2-8}
 &
  ChemBERTa-2 &
  RoBERTa &
  SMILES &
  Atom-level/Motif-level &
  MLM/MPR &
  Molecule Property Prediction & \cite{ahmad_chemberta-2_2022}
   \\\cmidrule(l){2-8}
 &
  MG-BERT &
  BERT &
  SMILES + Adjancency &
  Atom-level &
  MLM &
  Molecule Property Prediction & \cite{zhang_mg-bert_2021}
   \\\cmidrule(l){2-8}
 &
  X-Mol &
  \begin{tabular}[c]{@{}c@{}}X-Mol\\ (shared-layer \\ encoder-decoder)\end{tabular} &
  SMILES &
  Motif-level &
  \begin{tabular}[c]{@{}c@{}}ATG \\ (Representation Translation)\end{tabular} &
  \begin{tabular}[c]{@{}c@{}}Molecule Property Prediction\\ Yield Prediction\\ Drug-drug Interaction\\ Molecule Generation\end{tabular} & \cite{xue_x-mol_2022}
   \\\cmidrule(l){2-8}
 &
  ChemFormer &
  BART &
  SMILEs &
  Atom-level &
  \begin{tabular}[c]{@{}c@{}}MLM/ATG\\ (Representation Translation)\end{tabular} &
  \begin{tabular}[c]{@{}c@{}}Molecule Property Prediction\\ Molecule Generation\end{tabular} & \cite{irwin_chemformer_2022}
   \\\cmidrule(l){2-8}
 &
  BARTSmiles &
  BART &
  SMILES &
  Motif-level &
  MLM &
  \begin{tabular}[c]{@{}c@{}}Molecule Property Prediction\\ Reaction Prediction\end{tabular} & \cite{chilingaryan_bartsmiles_2022}
   \\\cmidrule(l){2-8}
 &
  SPT &
  GPT-3 &
  SMILES + Temperature &
  Character-level &
  MPR &
  Molecule Property Prediction & \cite{winter_smile_2022}
   \\\cmidrule(l){2-8}
 &
  T5 Chem &
  T5 &
  SMILES &
  Character-level &
  MLM &
  \begin{tabular}[c]{@{}c@{}}Reaction Type Classification\\ Reaction Yield Prediction\\ Reaction Prediction\end{tabular} & \cite{lu_unified_2022}
   \\\cmidrule(l){2-8}
 &
  MM-Deacon &
  Transformer &
  SMILES + IUPAC &
  Motif-level &
  XDC &
  \begin{tabular}[c]{@{}c@{}}Molecule Property Prediction\\ Cross-lingual Retrieval\\ Drug-Drug Interaction\end{tabular} & \cite{guo_multilingual_2022}
   \\\cmidrule(l){2-8}
 &
  Regression Transformer &
  XLNet &
  SELFIES &
  Atom-level + Numerical &
  \begin{tabular}[c]{@{}c@{}}ATG\\ (Property Prediction)\end{tabular} &
  Molecular Property Prediction & \cite{born_regression_2023}
   \\\cmidrule(l){2-8}
 &
  ChemGPT &
  GPT-3 &
  SELFIES &
  Atom-level &
  \begin{tabular}[c]{@{}c@{}}ATG\\ (Molecule Completion)\end{tabular} &
  NA & \cite{frey_neural_2023}
   \\\cmidrule(l){2-8}
 &
  MolFormer &
  RoFormer &
  SMILES &
  Atom-level &
  MLM &
  Molecule Property Prediction & \cite{wu_molformer_2023} 
     \\\cmidrule(l){2-8}
&
  Selformer &
  RoBERTa &
  SELFIES &
  Motif-level &
  MLM &
  Molecule Property Prediction & \cite{yuksel_selformer_2023}
   \\\midrule
\multicolumn{1}{c}{\multirow{9}{*}{\rotatebox[origin=c]{90}{\textbf{Multi-Domain}}}} &
  KV-PLM &
  BERT &
  SMILES, Text &
  Motif-level &
  MLM &
  \begin{tabular}[c]{@{}c@{}}Molecule Property Prediction\\ Reaction Type Classification\\ Molecule Captioning\\ De novo Molecule Generation\end{tabular} & \cite{zeng_deep-learning_2022}
   \\\cmidrule(l){2-8}
\multicolumn{1}{c}{} &
  MolT5 &
  T5 &
  SMILES, Text &
  Character-level &
  MLM &
  \begin{tabular}[c]{@{}c@{}}Molecule Captioning\\ De novo Molecule Generation\end{tabular} & \cite{edwards_translation_2022}
   \\\cmidrule(l){2-8}
\multicolumn{1}{c}{} &
  PrefixMol &
  GPT-3 &
  \begin{tabular}[c]{@{}c@{}}Property \\ Prefix Embedding\end{tabular} &
  Motif-level &
  \begin{tabular}[c]{@{}c@{}}ATG\\ (De Novo Molecule Generation)\end{tabular} &
  De Novo Molecule Generation & \cite{gao_prefixmol_2023}
   \\\cmidrule(l){2-8}
\multicolumn{1}{c}{} &
  BioT5 &
  T5 &
  SELFIES, Text, FASTA &
  Motif-level &
  \begin{tabular}[c]{@{}c@{}}MLM+ATG\\ (Molecule Captioning,\\De Novo Molecule Generation)\end{tabular} &
  \begin{tabular}[c]{@{}c@{}}Molecule Property Prediction\\ Molecule Captioning\\ De novo Molecule Generation\\ Drug-drug Interaction\\ Protein-Protein Interaction\\ Protein Property Prediction\end{tabular} & \cite{pei_biot5_2023}
   \\\cmidrule(l){2-8}
\multicolumn{1}{c}{} &
  MolGen &
  BART &
  SELFIES &
  Motif-level &
  MLM + Prefix Tuning &
  \begin{tabular}[c]{@{}c@{}}De novo Molecule Generation \\ Molecule Optimization\end{tabular}
   & \cite{fang_domain-agnostic_2023}
   \\\cmidrule(l){2-8}
\multicolumn{1}{c}{} &
  MolXPT &
  GPT2 &
  SMILES, Text &
  Atom-level &
  \begin{tabular}[c]{@{}c@{}}ATG\\ (Text Completion, Molecule Completion,\\ Molecule Captioning)\end{tabular} &
  \begin{tabular}[c]{@{}c@{}}Molecule Property Prediction\\ Molecule Captioning\\ De novo Molecule Generation\end{tabular} & \cite{liu_molxpt_2023}
   \\\cmidrule(l){2-8}
\multicolumn{1}{c}{} &
  Text+Chem T5 &
  T5 &
  SMILES, Text &
  Motif-level &
  MLM &
  \begin{tabular}[c]{@{}c@{}}Reaction Prediction\\ Molecule Captioning\\ De novo Molecule Generation\\ Paragraph to Action\end{tabular} & \cite{christofidellis_unifying_2023}
   \\\cmidrule(l){2-8}
\multicolumn{1}{c}{} &
  nach0 &
  T5 &
  SMILES, Text &
  Atom-level &
  \begin{tabular}[c]{@{}c@{}}ATG\\ (Molecule Property Prediction\\ Reaction Prediction\\ Molecule Captioning\\ De novo Molecule Generation)\end{tabular} &
  \begin{tabular}[c]{@{}c@{}}Molecule Property Prediction\\ Reaction Prediction\\ Molecule Captioning\\ De novo Molecule Generation\end{tabular} & \cite{livne_nach0_2023}
   \\\cmidrule(l){2-8}
\multicolumn{1}{c}{} &
  DrugGPT &
  GPT-2 &
  SMILES, Text, FASTA &
  Motif-level &
  \begin{tabular}[c]{@{}c@{}}ATG\\ (Molecule Completion)\end{tabular} &
  Drug Discovery & \cite{li_druggpt_2023}
   \\\midrule
\multirow{9}{*}{\rotatebox[origin=c]{90}{\textbf{Multi-modal}}} &
  Text2Mol &
  SciBERT &
  Graph, Text &
  Atom-level &
  XMC &
  Cross-modal Retrieval & \cite{edwards_text2mol_2021}
   \\\cmidrule(l){2-8}
 &
  DMP &
  RoBERTa &
  Graph, SMILES &
  Atom-level &
  MLM + XMC &
  \begin{tabular}[c]{@{}c@{}}Molecule Property Prediction\\ Reaction Prediction\end{tabular} & \cite{zhu_dual-view_2021}
   \\\cmidrule(l){2-8}
 &
  MoMu &
  SciBERT &
  Graph, Text &
  Atom-level &
  XMC &
  \begin{tabular}[c]{@{}c@{}}Molecule Property Prediction\\ Molecule Captioning\\ De novo Molecule Generataion\end{tabular} & \cite{su_molecular_2022}
   \\\cmidrule(l){2-8}
 &
  MolCA &
  Galactica/MolT5 &
  SMILES, Graph, Text &
  Motif-level &
  \begin{tabular}[c]{@{}c@{}} XMC + ATG\\ (Molecule Captioning)\end{tabular} &
  \begin{tabular}[c]{@{}c@{}}Cross-Modal Retrieval\\ Molecule Captioning\\ IUPCA Name Prediction\end{tabular} & \cite{liu_molca_2023}
   \\\cmidrule(l){2-8}
 &
  MolSTM &
  \begin{tabular}[c]{@{}c@{}}Chemformer\\ SciBERT\end{tabular} &
  SMILES, Graph, Text &
  Atom-level &
  XMC &
  \begin{tabular}[c]{@{}c@{}}Molecule Property Prediction\\ Cross-modal Retrieval\\ Molecule Generation\end{tabular} & \cite{liu_multi-modal_2023}
   \\\cmidrule(l){2-8}
 &
  GIT-Mol &
  SciBERT &
  \begin{tabular}[c]{@{}c@{}}Graph, Image, Text,\\ SMILES\end{tabular} &
  Character-level &
  XMC &
  Molecule Property Prediction & \cite{liu_git-mol_2023}
   \\\cmidrule(l){2-8}
 &
  GIMLET &
  T5 &
  Graph, Text &
  Atom-level &
  \begin{tabular}[c]{@{}c@{}}ATG\\ (Molecule Property Prediction)\end{tabular} &
  \begin{tabular}[c]{@{}c@{}}Molecule Property Prediction\\ (Zero-shot, few-shot)\end{tabular} &
   \cite{zhao_gimlet_2023} \\\cmidrule(l){2-8}
 &
  UniMap &
  RoBERTa &
  SMILES, Graph &
  Motif-level &
  MLM + XMC &
  \begin{tabular}[c]{@{}c@{}}Molecule Property Prediction\\ Drug-drug Interaction\end{tabular} & \cite{feng_unimap_2023}
   \\\cmidrule(l){2-8}
 &
  Memo &
  RoBERTa &
  \begin{tabular}[c]{@{}c@{}}Graph(2D+3D),\\ SMILES, Fingerprints\end{tabular} &
  Motif-level &
  MLM + XMC &
  Molecule Property Prediction &
  \cite{zhu_featurizations_2022}
  \\ \midrule
\end{tabular}%
}
\caption{Overview of Existing Approaches: The table columns on \textbf{Backbone}, \textbf{Tokenization Methods},and \textbf{Pretraining Objectives} specifically address the corresponding aspects related to chemical sequences in multi-domain and multi-modal approaches, including objectives for text encoders and alignment strategies in multi-modal settings.}
\label{tab:methods}
\end{table*}

\begin{figure*}
    \centering
    \begin{subfigure}{0.3\textwidth}
  \includegraphics[width=\linewidth]{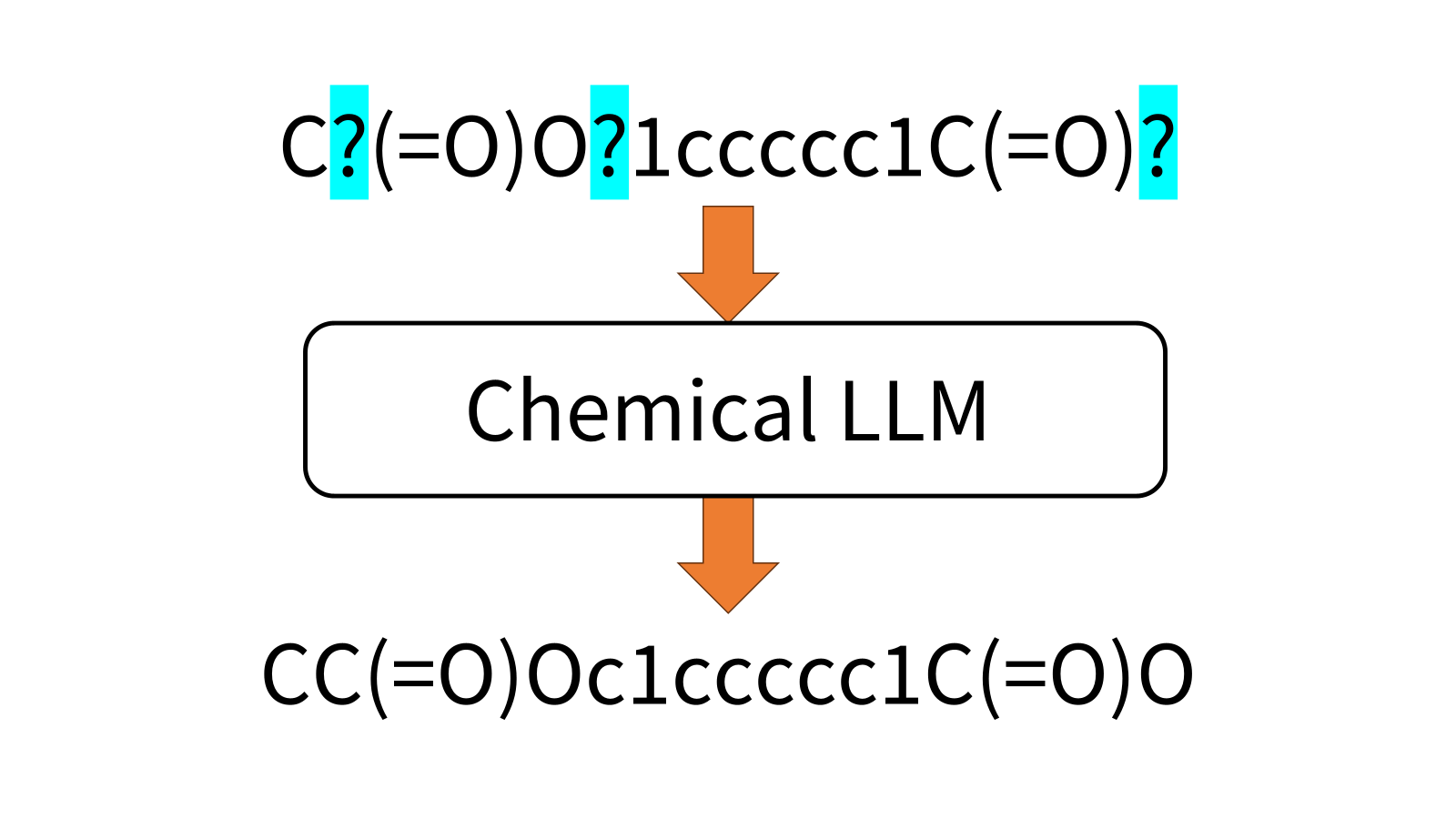}
  \caption{Masked Language Modelling Objective}
  \label{fig:MLM}
\end{subfigure}
\hfill 
\begin{subfigure}{0.3\textwidth}
  \includegraphics[width=\linewidth]{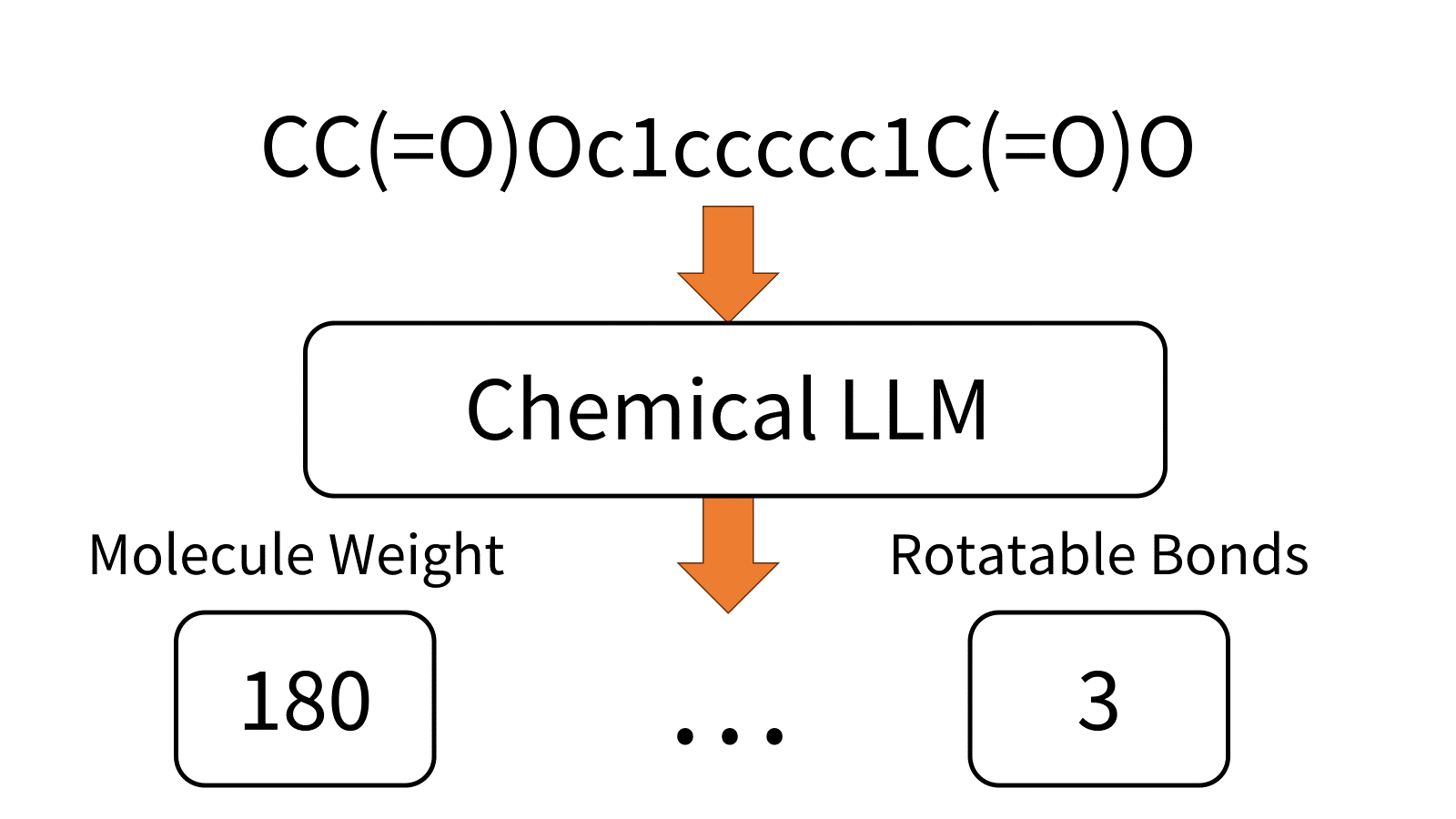}
  \caption{Molecule Property Prediction Objective}
  \label{fig:MPP}
\end{subfigure}
\hfill
\begin{subfigure}{0.3\textwidth}
    \includegraphics[width=\linewidth]{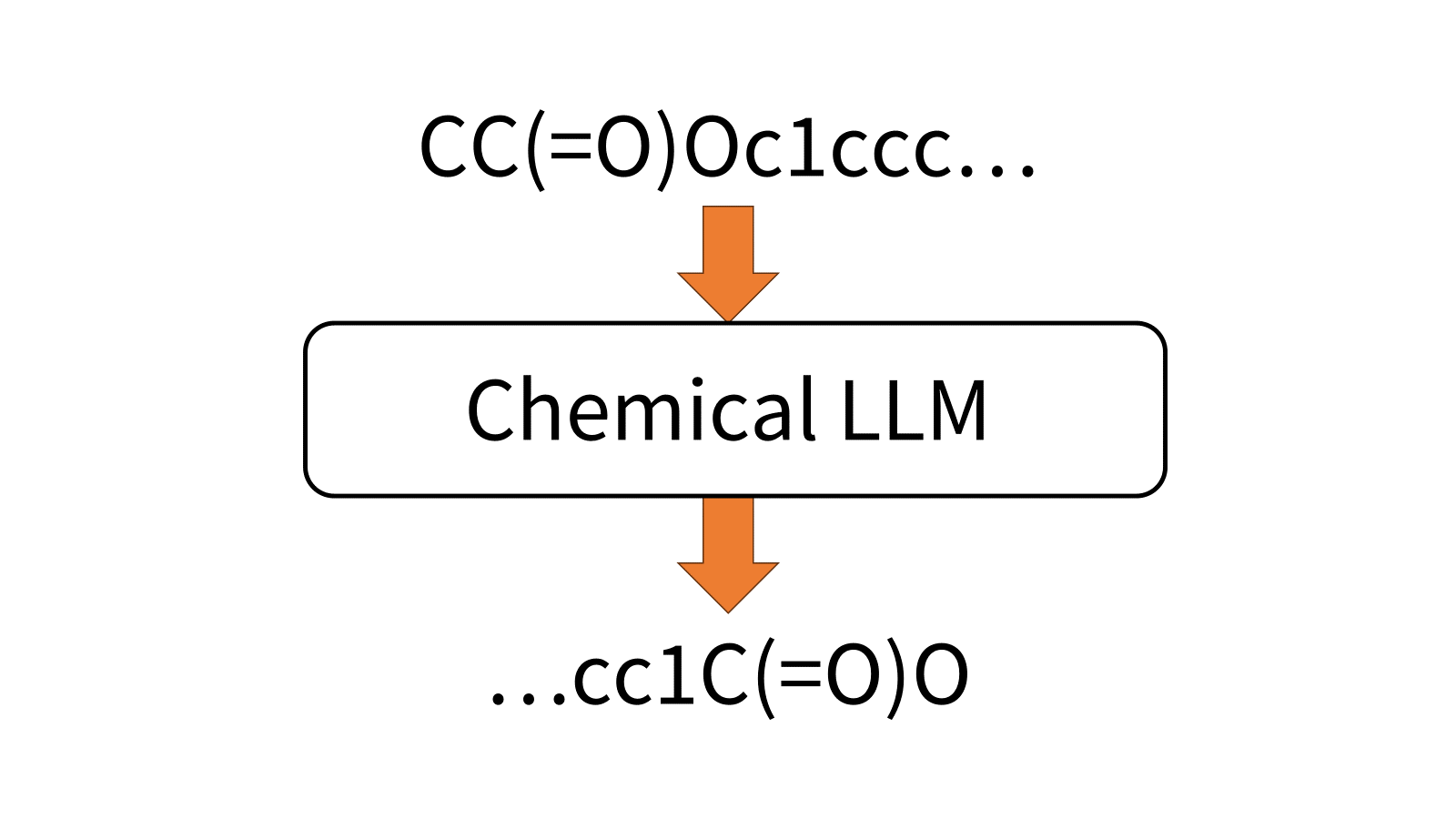}
    \caption{Autoregressive Token Generation Objective}
    \label{fig:ATG}
\end{subfigure}
\caption{Language Modelling Objectives}
\label{fig:pretrain}
\end{figure*}

\subsection{Language Modelling Objectives}
In this subsection, we review three key pretraining objectives: Masked Language Modeling (MLM), which directly applies to chemical LLMs; Molecule Property Prediction (MPP), specific to chemical domains; and Autoregressive Token Generation (ATG), adapted with chemistry-specific tasks for enhanced relevance in chemical LLMs. A graphical illustration for these objectives is shown in Figure~\ref{fig:pretrain}.

\paragraph{\textbf{Masked Language Modelling (MLM)}} is a prevalent pretraining objective for LLMs. It randomly substitutes tokens in the input sequence with a special ``[Mask]'' token or another arbitrary token from the vocabulary. The models are then trained to predict these masked tokens based on the surrounding context. For chemical LLMs, MLM is conducted on molecular sequential representations such as SMILES or SELFIES for single-domain approaches, and on wrapped sentences for multi-domain and multi-modal approaches. We define this objective as:
\begin{equation}
    \resizebox{.91\linewidth}{!}{$
            \displaystyle
\mathcal{L}_{\text{MLM}} = -\mathbb{E}_{\mathcal{S} \in \mathcal{D}} \left[ \sum_{\mathcal{S}' \in m(\mathcal{S})} \log p(\mathcal{S}' | \mathcal{S} \setminus m(\mathcal{S}')) \right]
$}
\end{equation} 
where \(m(\mathcal{S})\) represents the masked tokens in input sequence \(\mathcal{S}\). In single-domain approaches, SMILES-BERT \cite{wang_smiles-bert_2019}, ChemBERTa \cite{chithrananda_chemberta_2020}, MG-BERT \cite{zhang_mg-bert_2021}, Molformer \cite{ross_large-scale_2022}, Selformer \cite{yuksel_selformer_2023}, T5 Chem \cite{lu_unified_2022}, Chemformer \cite{irwin_chemformer_2022} and BARTSmiles \cite{chilingaryan_bartsmiles_2022} employ this objective on SMILES or SELFIES directly. MG-BERT \cite{zhang_mg-bert_2021} also enhances its input by incorporating graph adjacency knowledge, ensuring that attention calculations are confined to neighboring atoms only. Beyond single-domain, MLM extends to multi-domain and multi-modal approaches.  KV-PLM \cite{zeng_deep-learning_2022}, BioT5 \cite{pei_biot5_2023}, MultiTask Text+Chem T5 \cite{christofidellis_unifying_2023} perform MLM on wrapped sentences, DMP \cite{zhu_dual-view_2021}, Memo \cite{zhu_featurizations_2022}, UniMap \cite{feng_unimap_2023} leverage this objective for pretraining text encoders.

\paragraph{\textbf{Molecular Property Prediction (MPP)}} objective pretrains chemical LLMs to predict molecular properties given molecular sequential representations. This technique generates properties using cheminformatics tools such as RDKit, eliminating the requirement for manual labelling. These properties normally reflect the intrinsic semantics of molecules and typically include molecular weight, rotatable bond count, topological polar surface area, and the fraction of carbon atoms that are SP3 hybridized.
It can be formally defined as
\begin{equation}
       \mathcal{L}_{\text{MPP}}=-\mathbb{E}_{\mathcal{M} \in \mathcal{D}} \log p(\mathcal{P} | \mathcal{M})
\end{equation}
where \(\mathcal{P}\) is the set of calculated virtual properties for molecule \(\mathcal{M}\).
ChemBERTa-2 \cite{ahmad_chemberta-2_2022} conducts both MLM and MPP pretraining on a subset of PubChem \cite{chithrananda_chemberta_2020}, uncovering that MPP yields better converged performance albeit at the expense of increased training time. SPT \cite{winter_smile_2022} also utilizes the MPP at the early stage of pretraining and progresses to a laboratory-verified dataset to amplify performance. 
However, due to the design of MPP, MPP-based models are primarily used for molecular representation learning, as they cannot naturally generate tokens apart from properties. 

\paragraph{\textbf{Autoregressive Token Generation (ATG)}} refers to the scheme of generating the next token based on previous tokens. This objective can be formally defined as
\begin{equation}
   \mathcal{L}_{\text{ATG}}=-\mathbb{E}_{\mathcal{S} \in \mathcal{D}} \mathbb{E}_{t \in \mathcal{S}} p(t_i | t_0, t_1, ..., t_{i-1})
\end{equation}
where \(t_i\) stands for the \(i\)th token current predictin based on previous tokens \(t_0,...,t_{i-1}\) of the same sequence \(\mathcal{S}\). 
Data for downstream tasks, transformed into wrapped sentences for ATG pretraining, helps chemical LLMs adapt seamlessly. Given the significant overlap between ATG and downstream tasks, this section introduces pretraining-specific ATG tasks, with more downstream tasks usable for ATG pretraining available in Section~\ref{sec:application}.

\textbf{Molecule Completion} is the most straightforward approach wherein models are trained to complete the sequential representation of molecules. ChemGPT \cite{frey_neural_2023}, MolXPT \cite{liu_molxpt_2023}, and DrugGPT \cite{li_druggpt_2023} utilize this task to discern the underlying semantics of molecular sequential representations.

\textbf{Representation Translation} involves generating an alternative, viable representation from a given input sequence. Pioneering examples include X-mol \cite{xue_x-mol_2022} and Chemformer \cite{irwin_chemformer_2022}, which predict the canonical SMILES from a masked alternative SMILES representation.

\subsection{Cross-modal Objective}

\textbf{Cross-modal Contrastive (XMC)} learning maximizes the similarity within positive example pairs, while simultaneously emphasizes the distinction of negative pair examples.
In the context of chemistry-related multi-modal contrastive learning, representations of the same molecule, when presented across different modalities, are classified as positive pairs. Conversely, representations stemming from different molecules are designated as negative pairs, regardless of whether they occur within the same modality.

In the domain of contrastive learning, Info-NCE(Noise-Contrastive Estimation) loss \cite{oord_representation_2019} is a common training objective. It is defined as
\begin{equation}
\resizebox{.91\linewidth}{!}{$
    \displaystyle
    \mathcal{L}_{\text{Info-NCE}} = -\mathbb{E} \left[
    \sum_{M_x, M_y \in M} \log \frac{\exp(\text{sim}z_i^{M_x}, z_i^{M_y}) / \tau)}{\sum_{k=1}^{K} \exp(\text{sim}(z_i^{M_x}, z_i^{M_y}) / \tau)} \right]$}
\end{equation}
where \(M_x, M_y\) are two modalities of the overall modalities set \(M\) and \(z_i^{M_x}\) represents the hidden representation of \(i\)th example in modality \(M_x\).
MM-Deacon \cite{guo_multilingual_2022} applies Info-NCE objective on multi-lingual representations for molecules. Momu \cite{su_molecular_2022}, MOCO \cite{zhu_improving_2022}, MolCA \cite{liu_molca_2023}, MolSTM \cite{liu_multi-modal_2023} deploy Info-NCE loss to align representations from different modalities. 

Contrastive learning can also be employed by predicting whether two representations correspond to the same molecule. DMP \cite{zhu_dual-view_2021} enhances prediction accuracy by directly maximizing the cosine similarity between encodings of molecule graph and SMILES. Text2mol \cite{edwards_text2mol_2021} employs negative sampling loss. MolCA \cite{liu_molca_2023}, GIT-Mol \cite{liu_git-mol_2023}, and UniMap \cite{feng_unimap_2023} introduce additional prediction layers for contrastive learning, which are subsequently removed after pretraining. It's important to note that current cross-modal contrastive learning in chemical LLMs directly adapts from general contrastive learning approaches, potentially overlooking domain-specific nuances. Recently, the introduction of the reaction-aware contrastive learning framework PMSR \cite{jiang2023learning} proposes incorporating enhanced chemical knowledge, offering potential advancements for future multi-modal contrastive learning strategies.

\section{Applications}
\label{sec:application} 

\subsection{LLMs as Chatbots}
In multi-domain and multi-modal frameworks, pretraining with general text equips chemical LLMs with the ability to comprehend and respond to chemistry-related inquiries in textual format. This endows them with chatbot-like functionalities for downstream chemical tasks like MolT5 \cite{edwards_translation_2022}, BioT5 \cite{pei_biot5_2023}, MolXPT \cite{liu_molxpt_2023}, ChemGPT \cite{frey_neural_2023}, nach0 \cite{livne_nach0_2023}, and DrugGPT \cite{li_druggpt_2023}. These advanced systems enable users to engage in an intuitive dialogue format, posing inquiries in plain text and receiving detailed, contextually relevant responses, also in an accessible text format. 

A chatbot-like chemical LLM can also perform much more complex tasks involving understanding and reasoning.
\textbf{Experimental Action Extraction} task requires LLMs to extract specific experimental actions from verbose and detailed descriptions. Several methodologies have been developed for this task, as exemplified by the works of \cite{vaucher_automated_2020}, \cite{christofidellis_unifying_2023}, and \cite{vaskevicius_generative_2023}. Notably, \cite{vaucher_inferring_2021} advanced this field further by proposing a method to generate a series of experimental actions only given a specific reaction.

\textbf{Automated Laboratories} task enables researchers to input real-world queries, upon which the LLMs autonomously search scientific literature online, extract pivotal information, deduce practical synthesis routes, and execute experiments, all with minimal human intervention. Forefront initiatives like ChemCrow \cite{bran_chemcrow_2023}, Coscientist \cite{boiko_autonomous_2023}, and GPT-Lab \cite{qin_gpt-lab_2023} are spearheading this technological advancement in laboratory automation.

\subsection{LLMs as In-context Learners}
LLMs also possess an innate capability for learning directly from conversation-based interactions, an approach requiring no alterations to their model weights, known as in-context learning (ICL). \cite{guo_what_2023} performed a comprehensive assessment of LLMs on chemical tasks utilizing ICL. Their experimental results underscored that in-context learnt generalist LLMs could perform on par with chemistry-pretrained models in chemical tasks, thereby casting new light on the application of LLMs in the realm of chemistry. The effectiveness of in-context learning (ICL) is notably influenced by the style of prompts used, with simpler prompts often resulting in diminished performance, as evidenced in \cite{castro_nascimento_large_2023}. A well-constructed prompt in ICL should ideally comprise a general role-setting background, a concise task-specific introduction, relevant ICL examples, and a precise question as shown in Figure~\ref{fig:llmsinchemistry}. 
A well-designed prompt can even perform well without any ICL examples provided, such as those conducted by Synergpt \cite{edwards_synergpt_2023} and MolReGPT \cite{li_empowering_2023}. This achievement in zero-shot settings underscores the remarkable capabilities of Large Language Models (LLMs) in adapting to new tasks.

\subsection{LLMs as Representation Learners}
Similar to other pretraining models for representation learning, as discussed in \cite{xia_systematic_2023} and \cite{guo_graph-based_2023}, chemical LLMs are also adept at encoding molecular structures into latent spaces for training downstream models. A prevalent method involves using the [CLS] token representation at the beginning of sequences as a global molecular representation \cite{wang_smiles-bert_2019}, or employing shallow neural networks across all output tokens \cite{chithrananda_chemberta_2020}. In MG-BERT \cite{zhang_mg-bert_2021}, a virtual token that connects to all other tokens is included in the input sequence, and its representation is used to capture the global molecular structure for the input sequence. 
Additional task-specific modules are usually applied to encoded representations for downstream tasks. We list the common downstream applications as follows and comprehensive applications for each reviewed approach are shown in Table~\ref{tab:methods}: 
\begin{itemize}
    \item \textbf{Molecule Property Prediction} predicts properties with substantial industrial impact given a molecule that cannot be calculated from cheminformatic tools based on chemical sequences directly, such as blood-brain barrier permeability (BBBP) and lipophilicity.
    \item \textbf{Reaction Type/Yield Prediction} categorizing chemical reactions into specific types or predict productivity.
    \item \textbf{Reaction Prediction} encompasses three critical components: forward product prediction, single-step retrosynthesis, and reagent suggestion. Forward product prediction aims at forecasting the potential products of a given set of reactants. Single-step Retrosynthesis involves deducing feasible reactants from a desired product. Reagent suggestion focuses on identifying suitable reagents that facilitate a desired reaction.
    \item \textbf{Molecule Captioning \& De Novo Molecule Generation} are two novel tasks proposed in \cite{edwards_translation_2022}. Molecule captioning involves translating molecular representations into precise, textual descriptions and de novo molecule design presents the reverse challenge: creating novel molecules from textual descriptions.
    \item \textbf{Molecule Optimization} involves refining molecules to augment properties such as the Quantitative Estimate of Druglikeness (QED), lipophilicity and so on.
\end{itemize}

\section{Conclusion \& Future Directions}


To sum up, this survey offers a thorough exploration of the existing strategies for integrating LLMs into chemistry, covering the spectrum from input representation, through pretraining objectives, to diverse and unique applications. However, despite their rapid evolution, they remain in the nascent stages of development, indicating substantial room for growth and enhancement. The following future directions are pivotal for advancing the field:

\paragraph{Further Integration with Chemical Knowledge} Current chemical Large Language Models (LLMs) grapple with a limited grasp of the chemical universe, notably in retrosynthesis. The often-utilized USPTO\_50k dataset \cite{schneider_whats_2016}, with its 50,000 entries, barely scratches the surface of the vast and complex world of chemical retrosynthesis. This limitation significantly hampers the models' ability to comprehend and predict chemical retrosynthesis accurately. 
Additionally, as chemistry evolves with quantum mechanics into quantum chemistry, chemical LLMs are still rooted in conventional theories. This gap underscores the pressing need for these models to integrate more deeply with advanced chemical knowledge, particularly from quantum chemistry, to stay relevant and effective in modern chemical research.
\paragraph{Continual Learning} Once deployed, chemical LLMs encounter new knowledge incrementally, highlighting the necessity for continual learning. The requirement for continual learning is more urgent in applications like chemical reaction prediction due to reasons like variable synthesis routes, uncertain reaction conditions, etc. The Continual learning approach allows LLMs to adapt to fresh data from downstream tasks without forgetting previously acquired information, keeping them relevant and effective amidst the fast-paced evolution of chemical synthesis. 
\paragraph{Interpretability} LLMs often face criticism for their opaque, "black-box" nature, which obscures the rationale behind their outputs, rendering the results less interpretable to humans. The LLM4SD study \cite{zheng2023large} suggests leveraging LLMs for feature extraction, followed by the application of interpretable machine learning models, such as random forests or linear classifiers, on these features. Furthermore, the concept of Chain-of-Thought (CoT) prompting \cite{wei2022chain} has been introduced to prompt LLMs to articulate more intermediate reasoning steps, thereby enhancing their interpretability without compromising—and potentially even improving—their performance. Despite these advancements, the interpretability of chemical LLMs remains an unresolved challenge, presenting a valuable avenue for future research.

\bibliographystyle{named}
\bibliography{references}

\begin{thebibliography}{}

\bibitem[\protect\citeauthoryear{Ahmad \bgroup \em et al.\egroup }{2022}]{ahmad_chemberta-2_2022}
W.~Ahmad, et al.
\newblock {ChemBERTa}-2: {Towards} {Chemical} {Foundation} {Models}, 2022.

\bibitem[\protect\citeauthoryear{Axelrod and G{\'o}mez-Bombarelli}{2022}]{axelrod2022geom}
S.~Axelrod, et al.
\newblock Geom, energy-annotated molecular conformations for property prediction and molecular generation.
\newblock {\em Sci. Data}, 2022.

\bibitem[\protect\citeauthoryear{Boiko \bgroup \em et al.\egroup }{2023}]{boiko_autonomous_2023}
D.~A. Boiko, et al.
\newblock Autonomous chemical research with large language models.
\newblock {\em Nature}, 2023.

\bibitem[\protect\citeauthoryear{Born and Manica}{2023}]{born_regression_2023}
J.~Born, et al.
\newblock Regression {Transformer} enables concurrent sequence regression and generation for molecular language modelling.
\newblock {\em Nat. Mach. Intell}, 2023.

\bibitem[\protect\citeauthoryear{Bran \bgroup \em et al.\egroup }{2023}]{bran_chemcrow_2023}
A.~M. Bran, et al.
\newblock {ChemCrow}: {Augmenting} large-language models with chemistry tools, 2023.

\bibitem[\protect\citeauthoryear{Castro~Nascimento and Pimentel}{2023}]{castro_nascimento_large_2023}
C.~M. Castro~Nascimento, et al.
\newblock Do {Large} {Language} {Models} {Understand} {Chemistry}? {A} {Conversation} with {ChatGPT}.
\newblock {\em J. Chem. Inf. Model.}, 2023.

\bibitem[\protect\citeauthoryear{Chilingaryan \bgroup \em et al.\egroup }{2022}]{chilingaryan_bartsmiles_2022}
G.~Chilingaryan, et al.
\newblock {BARTSmiles}: {Generative} {Masked} {Language} {Models} for {Molecular} {Representations}, 2022.

\bibitem[\protect\citeauthoryear{Chithrananda \bgroup \em et al.\egroup }{2020}]{chithrananda_chemberta_2020}
S.~Chithrananda, et al.
\newblock {ChemBERTa}: {Large}-{Scale} {Self}-{Supervised} {Pretraining} for {Molecular} {Property} {Prediction}, 2020.

\bibitem[\protect\citeauthoryear{Christofidellis \bgroup \em et al.\egroup }{2023}]{christofidellis_unifying_2023}
D.~Christofidellis, et al.
\newblock Unifying {Molecular} and {Textual} {Representations} via {Multi}-task {Language} {Modelling}, 2023.

\bibitem[\protect\citeauthoryear{Degen \bgroup \em et al.\egroup }{2008}]{degen_art_2008}
J.~Degen, et al.
\newblock On the {Art} of {Compiling} and {Using} '{Drug}-{Like}' {Chemical} {Fragment} {Spaces}.
\newblock {\em ChemMedChem}, 2008.

\bibitem[\protect\citeauthoryear{Durant \bgroup \em et al.\egroup }{2002}]{durant_reoptimization_2002}
J.~L. Durant, et al.
\newblock Reoptimization of {MDL} {Keys} for {Use} in {Drug} {Discovery}.
\newblock {\em J. Chem. Inf. Comput.}, 2002.

\bibitem[\protect\citeauthoryear{Edwards \bgroup \em et al.\egroup }{2021}]{edwards_text2mol_2021}
C.~Edwards, et al.
\newblock {Text2Mol}: {Cross}-{Modal} {Molecule} {Retrieval} with {Natural} {Language} {Queries}.
\newblock In {\em EMNLP}, 2021.

\bibitem[\protect\citeauthoryear{Edwards \bgroup \em et al.\egroup }{2022}]{edwards_translation_2022}
C.~Edwards, et al.
\newblock Translation between molecules and natural language.
\newblock In {\em EMNLP}, 2022.

\bibitem[\protect\citeauthoryear{Edwards \bgroup \em et al.\egroup }{2023}]{edwards_synergpt_2023}
C.~Edwards, et al.
\newblock {SynerGPT}: {In}-{Context} {Learning} for {Personalized} {Drug} {Synergy} {Prediction} and {Drug} {Design}, 2023.

\bibitem[\protect\citeauthoryear{Fang \bgroup \em et al.\egroup }{2023a}]{fang_mol-instructions_2023}
Y.~Fang, et al.
\newblock Mol-{Instructions}: {A} {Large}-{Scale} {Biomolecular} {Instruction} {Dataset} for {Large} {Language} {Models}, 2023.

\bibitem[\protect\citeauthoryear{Fang \bgroup \em et al.\egroup }{2023b}]{fang_domain-agnostic_2023}
Y.~Fang, et al.
\newblock Domain-{Agnostic} {Molecular} {Generation} with {Self}-feedback, 2023.

\bibitem[\protect\citeauthoryear{Feng \bgroup \em et al.\egroup }{2023}]{feng_unimap_2023}
S.~Feng, et al.
\newblock {UniMAP}: {Universal} {SMILES}-{Graph} {Representation} {Learning}, 2023.

\bibitem[\protect\citeauthoryear{Frey \bgroup \em et al.\egroup }{2023}]{frey_neural_2023}
N.~C. Frey, et al.
\newblock Neural scaling of deep chemical models.
\newblock {\em Nat. Mach. Intell}, 2023.

\bibitem[\protect\citeauthoryear{Gage}{1994}]{gage_new_1994}
P.~Gage.
\newblock A new algorithm for data compression.
\newblock {\em The C Users Journal}, 1994.

\bibitem[\protect\citeauthoryear{Gao \bgroup \em et al.\egroup }{2023}]{gao_prefixmol_2023}
Z.~Gao, et al.
\newblock {PrefixMol}: {Target}- and {Chemistry}-aware {Molecule} {Design} via {Prefix} {Embedding}, 2023.

\bibitem[\protect\citeauthoryear{Guo \bgroup \em et al.\egroup }{2022}]{guo_multilingual_2022}
Z.~Guo, et al.
\newblock Multilingual molecular representation learning via contrastive pre-training.
\newblock In {\em ACL}, 2022.

\bibitem[\protect\citeauthoryear{Guo \bgroup \em et al.\egroup }{2023a}]{guo_what_2023}
T.~Guo, et al.
\newblock What can {Large} {Language} {Models} do in chemistry? {A} comprehensive benchmark on eight tasks, 2023.

\bibitem[\protect\citeauthoryear{Guo \bgroup \em et al.\egroup }{2023b}]{guo_graph-based_2023}
Z.~Guo, et al.
\newblock Graph-based molecular representation learning.
\newblock In {\em IJCAI}, 2023.

\bibitem[\protect\citeauthoryear{He \bgroup \em et al.\egroup }{2016}]{he2016deep}
K.~He, et al.
\newblock Deep residual learning for image recognition.
\newblock In {\em CVPR}, 2016.

\bibitem[\protect\citeauthoryear{Heller \bgroup \em et al.\egroup }{2013}]{heller_inchi_2013}
S.~Heller, et al.
\newblock {InChI} - the worldwide chemical structure identifier standard.
\newblock {\em J. Cheminformatics}, 2013.

\bibitem[\protect\citeauthoryear{Hopfinger \bgroup \em et al.\egroup }{1997}]{hopfinger1997construction}
A.~Hopfinger, et al.
\newblock Construction of 3d-qsar models using the 4d-qsar analysis formalism.
\newblock {\em J. Am. Chem. Soc.}, 1997.

\bibitem[\protect\citeauthoryear{Irwin \bgroup \em et al.\egroup }{2020}]{irwin_zinc20free_2020}
J.~J. Irwin, et al.
\newblock {ZINC20}—{A} {Free} {Ultralarge}-{Scale} {Chemical} {Database} for {Ligand} {Discovery}.
\newblock {\em J. Chem. Inf. Model.}, 2020.

\bibitem[\protect\citeauthoryear{Irwin \bgroup \em et al.\egroup }{2022}]{irwin_chemformer_2022}
R.~Irwin, et al.
\newblock Chemformer: a pre-trained transformer for computational chemistry.
\newblock {\em MLST}, 2022.

\bibitem[\protect\citeauthoryear{Jiang \bgroup \em et al.\egroup }{2023}]{jiang2023learning}
Y.~Jiang, et al.
\newblock Learning chemical rules of retrosynthesis with pre-training.
\newblock In {\em AAAI}, 2023.

\bibitem[\protect\citeauthoryear{Jin \bgroup \em et al.\egroup }{2018}]{jin_junction_2019}
W.~Jin, et al.
\newblock Junction tree variational autoencoder for molecular graph generation.
\newblock In {\em ICML}, 2018.

\bibitem[\protect\citeauthoryear{Kim \bgroup \em et al.\egroup }{2016}]{kim_pubchem_2016}
S.~Kim, et al.
\newblock {PubChem} {Substance} and {Compound} databases.
\newblock {\em Nucleic Acids Res.}, 2016.

\bibitem[\protect\citeauthoryear{Krenn \bgroup \em et al.\egroup }{2020}]{krenn_self-referencing_2020}
M.~Krenn, et al.
\newblock Self-referencing embedded strings ({SELFIES}): {A} 100\% robust molecular string representation.
\newblock {\em MLST}, 2020.

\bibitem[\protect\citeauthoryear{Li and Fourches}{2021}]{li_smiles_2021}
X.~Li, et al.
\newblock {SMILES} {Pair} {Encoding}: {A} {Data}-{Driven} {Substructure} {Tokenization} {Algorithm} for {Deep} {Learning}.
\newblock {\em J. Chem. Inf. Model}, 2021.

\bibitem[\protect\citeauthoryear{Li \bgroup \em et al.\egroup }{2023a}]{li_empowering_2023}
J.~Li, et al.
\newblock Empowering {Molecule} {Discovery} for {Molecule}-{Caption} {Translation} with {Large} {Language} {Models}: {A} {ChatGPT} {Perspective}, 2023.

\bibitem[\protect\citeauthoryear{Li \bgroup \em et al.\egroup }{2023b}]{li_druggpt_2023}
Y.~Li, et al.
\newblock {DrugGPT}: {A} {GPT}-based {Strategy} for {Designing} {Potential} {Ligands} {Targeting} {Specific} {Proteins}, 2023.

\bibitem[\protect\citeauthoryear{Liu \bgroup \em et al.\egroup }{2023a}]{liu_git-mol_2023}
P.~Liu, et al.
\newblock {GIT}-{Mol}: {A} {Multi}-modal {Large} {Language} {Model} for {Molecular} {Science} with {Graph}, {Image}, and {Text}, 2023.

\bibitem[\protect\citeauthoryear{Liu \bgroup \em et al.\egroup }{2023b}]{liu_multi-modal_2023}
S.~Liu, et al.
\newblock Multi-modal molecule structure--text model for text-based retrieval and editing.
\newblock {\em Nat. Mach. Intell}, 2023.

\bibitem[\protect\citeauthoryear{Liu \bgroup \em et al.\egroup }{2023c}]{liu_molxpt_2023}
Z.~Liu, et al.
\newblock {MolXPT}: {Wrapping} {Molecules} with {Text} for {Generative} {Pre}-training.
\newblock ACL, 2023.

\bibitem[\protect\citeauthoryear{Liu \bgroup \em et al.\egroup }{2023d}]{liu_molca_2023}
Z.~Liu, et al.
\newblock {M}ol{CA}: Molecular graph-language modeling with cross-modal projector and uni-modal adapter.
\newblock In {\em EMNLP}, 2023.

\bibitem[\protect\citeauthoryear{Livne \bgroup \em et al.\egroup }{2023}]{livne_nach0_2023}
M.~Livne, et al.
\newblock nach0: {Multimodal} {Natural} and {Chemical} {Languages} {Foundation} {Model}, 2023.

\bibitem[\protect\citeauthoryear{Lu and Zhang}{2022}]{lu_unified_2022}
J.~Lu, et al.
\newblock Unified deep learning model for multitask reaction predictions with explanation.
\newblock {\em J. Chem. Inf. Model}, 2022.

\bibitem[\protect\citeauthoryear{Oord \bgroup \em et al.\egroup }{2019}]{oord_representation_2019}
A.~v.~d. Oord, et al.
\newblock Representation {Learning} with {Contrastive} {Predictive} {Coding}, 2019.

\bibitem[\protect\citeauthoryear{Pei \bgroup \em et al.\egroup }{2023}]{pei_biot5_2023}
Q.~Pei, et al.
\newblock {B}io{T}5: Enriching cross-modal integration in biology with chemical knowledge and natural language associations.
\newblock In {\em EMNLP}, 2023.

\bibitem[\protect\citeauthoryear{Qin \bgroup \em et al.\egroup }{2023}]{qin_gpt-lab_2023}
X.~Qin, et al.
\newblock {GPT}-{Lab}: {Next} {Generation} {Of} {Optimal} {Chemistry} {Discovery} {By} {GPT} {Driven} {Robotic} {Lab}, 2023.

\bibitem[\protect\citeauthoryear{Raffel \bgroup \em et al.\egroup }{2020}]{c4}
C.~Raffel, et al.
\newblock Exploring the limits of transfer learning with a unified text-to-text transformer.
\newblock {\em JMLR}, 2020.

\bibitem[\protect\citeauthoryear{Rogers and Hahn}{2010}]{rogers_extended-connectivity_2010}
D.~Rogers, et al.
\newblock Extended-{Connectivity} {Fingerprints}.
\newblock {\em J. Chem. Inf. Model}, 2010.

\bibitem[\protect\citeauthoryear{Rong \bgroup \em et al.\egroup }{2020}]{rong_self-supervised_2020}
Y.~Rong, et al.
\newblock Self-supervised graph transformer on large-scale molecular data.
\newblock In {\em NeurIPS}, NIPS'20, 2020.

\bibitem[\protect\citeauthoryear{Ross \bgroup \em et al.\egroup }{2022}]{ross_large-scale_2022}
J.~Ross, et al.
\newblock Large-scale chemical language representations capture molecular structure and properties.
\newblock {\em Nat. Mach. Intell}, 2022.

\bibitem[\protect\citeauthoryear{Schneider \bgroup \em et al.\egroup }{2016}]{schneider_whats_2016}
N.~Schneider, et al.
\newblock What’s {What}: {The} ({Nearly}) {Definitive} {Guide} to {Reaction} {Role} {Assignment}.
\newblock {\em J. Chem. Inf. Model}, 2016.

\bibitem[\protect\citeauthoryear{Sch{\"u}tt \bgroup \em et al.\egroup }{2017}]{schutt2017schnet}
K.~Sch{\"u}tt, et al.
\newblock Schnet: A continuous-filter convolutional neural network for modeling quantum interactions.
\newblock In {\em NIPS}, 2017.

\bibitem[\protect\citeauthoryear{Schwaller \bgroup \em et al.\egroup }{2018}]{schwaller_found_2017}
P.~Schwaller, et al.
\newblock “found in translation”: predicting outcomes of complex organic chemistry reactions using neural sequence-to-sequence models.
\newblock {\em Chem. Sci}, 2018.

\bibitem[\protect\citeauthoryear{Schwaller \bgroup \em et al.\egroup }{2019}]{schwaller_molecular_2019}
P.~Schwaller, et al.
\newblock Molecular {Transformer}: {A} {Model} for {Uncertainty}-{Calibrated} {Chemical} {Reaction} {Prediction}.
\newblock {\em ACS Cent. Sci}, 2019.

\bibitem[\protect\citeauthoryear{Su \bgroup \em et al.\egroup }{2022}]{su_molecular_2022}
B.~Su, et al.
\newblock A {Molecular} {Multimodal} {Foundation} {Model} {Associating} {Molecule} {Graphs} with {Natural} {Language}, 2022.

\bibitem[\protect\citeauthoryear{Vaswani \bgroup \em et al.\egroup }{2017}]{vaswani2017attention}
A.~Vaswani, et al.
\newblock Attention is all you need.
\newblock In {\em NIPS}, 2017.

\bibitem[\protect\citeauthoryear{Vaucher \bgroup \em et al.\egroup }{2020}]{vaucher_automated_2020}
A.~C. Vaucher, et al.
\newblock Automated extraction of chemical synthesis actions from experimental procedures.
\newblock {\em Nat. Commun}, 2020.

\bibitem[\protect\citeauthoryear{Vaucher \bgroup \em et al.\egroup }{2021}]{vaucher_inferring_2021}
A.~C. Vaucher, et al.
\newblock Inferring experimental procedures from text-based representations of chemical reactions.
\newblock {\em Nat. Commun}, 2021.

\bibitem[\protect\citeauthoryear{Vaškevičius \bgroup \em et al.\egroup }{2023}]{vaskevicius_generative_2023}
M.~Vaškevičius, et al.
\newblock Generative {LLMs} in {Organic} {Chemistry}: {Transforming} {Esterification} {Reactions} into {Natural} {Language} {Procedures}.
\newblock {\em Appl. Sci.}, 2023.

\bibitem[\protect\citeauthoryear{Wang \bgroup \em et al.\egroup }{2019}]{wang_smiles-bert_2019}
S.~Wang, et al.
\newblock {SMILES}-{BERT}: {Large} {Scale} {Unsupervised} {Pre}-{Training} for {Molecular} {Property} {Prediction}.
\newblock BCB, 2019.

\bibitem[\protect\citeauthoryear{Wei \bgroup \em et al.\egroup }{2022}]{wei2022chain}
J.~Wei, et al.
\newblock Chain-of-thought prompting elicits reasoning in large language models.
\newblock {\em NeurIPS}, 2022.

\bibitem[\protect\citeauthoryear{Weininger}{1988}]{weininger_smiles_1988}
D.~Weininger.
\newblock {SMILES}, a chemical language and information system. 1. {Introduction} to methodology and encoding rules.
\newblock {\em J. Chem. Inf. Comput.}, 1988.

\bibitem[\protect\citeauthoryear{Winter \bgroup \em et al.\egroup }{2022}]{winter_smile_2022}
B.~Winter, et al.
\newblock A smile is all you need: predicting limiting activity coefficients from {SMILES} with natural language processing.
\newblock {\em Digital Discovery}, 2022.

\bibitem[\protect\citeauthoryear{Wu \bgroup \em et al.\egroup }{2023}]{wu_molformer_2023}
F.~Wu, et al.
\newblock Molformer: {Motif}-{Based} {Transformer} on {3D} {Heterogeneous} {Molecular} {Graphs}.
\newblock {\em AAAI}, 2023.

\bibitem[\protect\citeauthoryear{Xia \bgroup \em et al.\egroup }{2023}]{xia_systematic_2023}
J.~Xia, et al.
\newblock A systematic survey of chemical pre-trained models.
\newblock In {\em IJCAI}, 2023.

\bibitem[\protect\citeauthoryear{Xie \bgroup \em et al.\egroup }{2023}]{xie_self-supervised_2023}
A.~Xie, et al.
\newblock Self-supervised learning with chemistry-aware fragmentation for effective molecular property prediction.
\newblock {\em Brief. Bioinform}, 2023.

\bibitem[\protect\citeauthoryear{Xu \bgroup \em et al.\egroup }{2019}]{xu2019powerful}
K.~Xu, et al.
\newblock How powerful are graph neural networks?
\newblock In {\em ICLR}, 2019.

\bibitem[\protect\citeauthoryear{Xue \bgroup \em et al.\egroup }{2022}]{xue_x-mol_2022}
D.~Xue, et al.
\newblock X-{MOL}: large-scale pre-training for molecular understanding and diverse molecular analysis.
\newblock {\em Sci. Bull.}, 2022.

\bibitem[\protect\citeauthoryear{Yüksel \bgroup \em et al.\egroup }{2023}]{yuksel_selformer_2023}
A.~Yüksel, et al.
\newblock {SELFormer}: molecular representation learning via {SELFIES} language models.
\newblock {\em MLST}, 2023.

\bibitem[\protect\citeauthoryear{Zeng \bgroup \em et al.\egroup }{2022}]{zeng_deep-learning_2022}
Z.~Zeng, et al.
\newblock A deep-learning system bridging molecule structure and biomedical text with comprehension comparable to human professionals.
\newblock {\em Nat. Commun}, 2022.

\bibitem[\protect\citeauthoryear{Zhang \bgroup \em et al.\egroup }{2021a}]{zhang_mg-bert_2021}
X.-C. Zhang, et al.
\newblock {MG}-{BERT}: leveraging unsupervised atomic representation learning for molecular property prediction.
\newblock {\em Brief. Bioinform}, 2021.

\bibitem[\protect\citeauthoryear{Zhang \bgroup \em et al.\egroup }{2021b}]{zhang_motif-based_2021}
Z.~Zhang, et al.
\newblock Motif-based graph self-supervised learning for molecular property prediction.
\newblock {\em NeurIPS}, 2021.

\bibitem[\protect\citeauthoryear{Zhao \bgroup \em et al.\egroup }{2023}]{zhao_gimlet_2023}
H.~Zhao, et al.
\newblock {GIMLET}: A unified graph-text model for instruction-based molecule zero-shot learning.
\newblock In {\em NeurIPS}, 2023.

\bibitem[\protect\citeauthoryear{Zheng \bgroup \em et al.\egroup }{2023}]{zheng2023large}
Y.~Zheng, et al.
\newblock Large language models for scientific synthesis, inference and explanation, 2023.

\bibitem[\protect\citeauthoryear{Zhu \bgroup \em et al.\egroup }{2022}]{zhu_featurizations_2022}
Y.~Zhu, et al.
\newblock Featurizations matter: A multiview contrastive learning approach to molecular pretraining.
\newblock In {\em ICML AI for Science Workshop}, 2022.

\bibitem[\protect\citeauthoryear{Zhu \bgroup \em et al.\egroup }{2023}]{zhu_dual-view_2021}
J.~Zhu, et al.
\newblock Dual-view molecular pre-training.
\newblock In {\em SIGKDD}, 2023.

\end{thebibliography}

\end{document}